\documentclass[journal]{IEEEtran}

\usepackage{amsmath,amsfonts}
\usepackage{amsthm}
\usepackage{amssymb}
\usepackage{enumitem}
\usepackage{array, multirow, graphicx}
\usepackage[caption=false,font=normalsize,labelfont=sf,textfont=sf]{subfig}
\usepackage{mathtools}
\usepackage[T1]{fontenc}
\usepackage{url}

\newtheorem{corollary}{Corollary}
\newtheorem{req}{Requirement}

\theoremstyle{definition}
\newtheorem{definition}{Definition}

\begin{document}

\title{Foundational Requirements for Artificial General Intelligence: A Falsifiable Framework Based on Signal Prediction}

\author{\textbf{Matej \v{S}progar}\\University of Maribor, Faculty of Electrical Engineering and Computer Science, Maribor, Slovenia\\matej.sprogar@um.si}

\maketitle

\begin{abstract}
Grounded in the premise that high-level intelligence can emerge from low-level signal processing, we hypothesise a set of requirements necessary for artificial general intelligence. They characterise systems that learn by predicting spatially and temporally structured signals of initially unknown meaning, and draw on principles observed in cognitive neuroscience, from learning from an uninformed state to real-time liveness.

To enable empirical testing and rejection of the hypothesis, we introduce a testbed of transparent, reusable tests, one per requirement. Failing a test exposes a violation of a requirement, but passing is not conclusive: a transparent testbed cannot detect systems engineered around its tests, and such systems can pass it without being intelligent. Passing is therefore meaningful only for systems developed without knowledge of the tests, and none has yet been reported. The testbed thus serves as a necessary-condition filter rather than as proof of progress toward general intelligence. Its reference implementation is publicly available.
\end{abstract}


\section{Introduction}

Intelligence remains one of the deepest unresolved questions in science, as we still lack both a precise definition of intelligence and an understanding of its fundamental components \cite{Turing1950, Guo2026, PlatoAI2026}. Assuming there is such a thing as general intelligence, one major obstacle to creating artificial general intelligence (AGI) is the limited understanding of how biological brains operate. At the same time, modern artificial intelligence (AI) systems achieve impressive performance across a wide range of tasks \cite{StanfordAIIndex2025}. However, such performance is often achieved without a principled account of understanding and may not reflect broadly general capabilities \cite{Marcus2018, Mitchell2025, Shojaee2025}.

Until we discover the secret behind general intelligence, our models can only aim to simulate the observable behaviour it produces. As a result, modern AI conflates specialised competence with general intelligence \cite{BenderKoller2020, Bisk2020, Srivastava2023}. Systems optimised for a narrow class of tasks may perform extremely well within that domain while remaining brittle or ineffective elsewhere. 

The existing definitions and descriptions of intelligent behaviour are useless for AGI development because they are either vague, self-referential, or non-general. However valuable modern AI is, maybe AGI is not achievable by targeting a high-level behaviour on a predefined task. Maybe we should focus on the low-level behaviours necessary for the emergence of general intelligent behaviour. This perspective is motivated in part by observations from biological neural systems, whose operation is often described in terms of continuous streams of sensory signals rather than discrete symbolic\footnote{Symbolic representations introduce an additional complication known as the symbol grounding problem, in which symbolic tokens require grounding in perception or environmental interaction to acquire intrinsic semantic meaning \cite{Harnad1990}.} representations \cite{Friston2010, ChurchlandSejnowski2016}. Working directly with low-level signals offers a way to avoid assuming pre-existing semantic structure and allows us to propose testable, operational requirements for the emergence of intelligent behaviour.
 
In this paper, we identify requirements hypothesised to be necessary but not sufficient for general intelligence, and formalise them as architecture-agnostic criteria. Although each requirement can be satisfied individually by specialised systems, complexity arises from their interaction; consequently, the challenge lies in satisfying them simultaneously. We propose twelve requirements as a minimal set hypothesised to capture the essential conditions for general intelligence.

Because intelligence lacks a precise definition, the hypothesis cannot be directly verified and is therefore formulated in a falsifiable manner. To this end, we introduce the Artificial General Intelligence Testbed (AGITB), which evaluates whether a system under test satisfies all proposed requirements. Successfully passing the testbed does not, by itself, establish the presence of general intelligence. However, the existence of a non-intelligent system that satisfies all requirements would falsify the hypothesis. Because the tests are transparent, a system can also be engineered to pass them without satisfying the requirements they approximate; passing the AGITB is therefore meaningful only for systems that were not designed with knowledge of the individual tests. Failing it, by contrast, indicates that a system remains limited to narrow or specialised capabilities.

The originality of this work lies not in any individual component considered in isolation, but in the synthesis of minimal, low-level, architecture-independent behavioural requirements. Formalising these requirements enabled the construction of a transparent, reusable testbed for the empirical evaluation of experimental AGI systems, thereby addressing a longstanding methodological gap in the field.

\section{Relationship to Prior Research}

The present work intersects several established research traditions, including predictive processing and predictive coding in computational neuroscience \cite{Rao1999,Friston2010}, continual and sequence learning \cite{Hochreiter1997,Hawkins2005}, and formal theories of intelligence \cite{Hutter2005,Legg2007}. However, to our knowledge, no existing framework combines the specific set of objectives pursued here within a single operational, falsifiable system.

Conceptually, the framework is most closely aligned with predictive-processing accounts of cognition, in which intelligent behaviour emerges through continual prediction over temporally structured input streams \cite{Clark2013,Friston2010}. Related ideas also appear in sequence-memory approaches such as Hierarchical Temporal Memory \cite{Hawkins2005}, where prediction and temporal state evolution are treated as central computational primitives. Existing approaches of this kind, however, are primarily architectural or neurocomputational proposals rather than attempts to derive architecture-independent necessary conditions for general intelligence.

At a more abstract level, the work also relates to formal theories of intelligence, including universal intelligence and AIXI-style formulations \cite{Hutter2005,Legg2007}, which similarly seek principled characterisations of intelligence from minimal assumptions. Unlike the present work, however, these approaches do not operationalise their assumptions as architecture-independent, falsifiable criteria for empirical evaluation.

Our approach further differs from contemporary benchmark-oriented paradigms for evaluating intelligence, including task-generalisation benchmarks such as ARC \cite{Chollet2019}. Modern machine-learning evaluations are typically constructed around finite test distributions, with hidden or periodically refreshed datasets intended to limit optimisation toward the evaluation itself. By contrast, the proposed testbed does not rely on benchmark secrecy; instead, it introduces transparent, comprehensible tasks. Success depends on satisfying the underlying behavioural constraints across combinatorially large input spaces rather than on reproducing finite hidden test distributions. The price of transparency is that the tests can be studied and targeted: a system designed around them can pass without possessing the capabilities they are meant to probe (Section~\ref{sec:exploitation}).

\section{Background}

Contemporary artificial systems do not yet robustly exhibit behaviour indistinguishable from that of humans. One possible research direction towards that goal involves closer alignment with the computational principles observed in the human cortex, motivating interest in neuromorphic approaches such as spiking neural networks, which explicitly incorporate time-sensitive, event-driven computation \cite{Maass1997, Gerstner2002}. While such architectures do not, in themselves, constitute general intelligence, they highlight alternative design dimensions that remain relatively underexplored in conventional artificial neural networks.

In alignment with this biologically grounded perspective, we depart from symbolic, high-level behaviour and instead operate at the level of raw signals. While Turing \cite{Turing1950} was right to suggest that communication could serve as a basis for evaluating machine intelligence, natural language remains problematic as a test medium: it conveys human knowledge through symbols whose meanings are not intrinsically grounded in machines, as argued by Stevan Harnad \cite{Harnad1990}. As a result, symbolic evaluation can obscure whether a system has acquired grounded representations or merely exploits statistical regularities in linguistic form. Although the symbol-grounding problem is an old philosophical issue, it has regained prominence across cognitive science, neuroscience, and machine learning \cite{BenderKoller2020, Bisk2020, Gubelmann2024}.

\subsection{Biological grounding and non-symbolic evaluation}

To minimise assumptions about semantic grounding, we focus on systems that operate directly on the lowest-level binary signals, each represented as a structure of bits processed in parallel. A bit corresponds directly to the state of a measurable signal channel (such as a pixel, an audio band, or actuator feedback) and, as the smallest unit of information, does not in itself impose higher-level semantic structure. Binary signals also provide a simple abstraction of neural spiking events in biological systems.

This perspective is consistent with the functioning of the cerebral cortex, where cognition arises from the processing of temporally structured sensory spike trains rather than disembodied symbols. The “brain simulator” reply to John Searle’s Chinese Room argument \cite{Searle1980} suggests that a system reproducing the causal dynamics of cortical processing could, in principle, instantiate the same cognitive mechanisms \cite{Churchland1990}. The cortex represents at least one plausible route toward general intelligence.

The proposed AGI requirements are particularly well-suited for NeuroAI models and align with the principles of the embodied Turing Test \cite{Zador2023}, in which cognitive understanding is hypothesised to emerge from the integration of continuous sensory streams. The progression from raw-signal prediction to higher-level abstraction mirrors that from early perceptrons to contemporary large-scale models.

\subsection{Integrated requirements for intelligence}

A starting point for the AGI requirements is the \emph{ladder to human-comparable intelligence} \cite{Sprogar2018}, a framework that organises cognitive capabilities into progressively more demanding levels. Its first three rungs align naturally with the framework proposed here. However, the ladder's individual requirements can be met by specialised systems that lack general intelligence, which creates a risk of false positives during evaluation. Minimising such false positives, therefore, requires an all-or-nothing criterion: a system must satisfy \emph{all} requirements simultaneously rather than meeting them individually. In addition, the higher rungs of the ladder remain insufficiently specified for operational evaluation.

Consequently, failure on any single requirement is sufficient to rule out AGI, whereas satisfying all of them does not establish it. This complementarity allows individual requirements to remain simple in scope, while their joint satisfaction is intended to impose constraints that narrow or specialised systems cannot easily meet.

\subsection{Learning through prediction}

An intelligent system learns to recognise the structure of input signals originating from the external world. The learning process cannot rely solely on externally defined performance metrics, such as accuracy or mean-squared error, because predefined quantitative metrics struggle to capture the richness and variability of real-world structure.

This limitation is reflected in many existing approaches to evaluating AGI, which remain constrained by metrics that inadequately capture the breadth of general intelligence. Contemporary systems already surpass humans on many standardised tasks---for example, playing chess---illustrating how narrowly defined benchmarks can be optimised without necessarily yielding broadly general capabilities. Metric-based evaluation can therefore encourage systems engineered to maximise specific scores rather than those exhibiting general intelligence.

Achieving generality, therefore, requires a learning process driven by an internally defined objective that does not rely on externally supplied semantics. One such objective is the ability to predict the future sensory signals the system will experience. Prediction at the level of sensory signals provides a foundation for higher-level expectations about the external world to emerge without requiring predefined semantic representations. Biological intelligence is often characterised as extracting structure from sensory data in order to anticipate future states \cite{Hawkins2005, Clark2013}.

Accordingly, we require a generally intelligent system to predict future inputs from past observations. Prediction alone does not constitute intelligence, but the ability to anticipate future sensory inputs is widely considered a fundamental capability underlying perception, learning, and action \cite{Marais2025, Darriba2021}.

\subsection{Spatial and temporal structure of input signals}
\label{sec:structure}

In the operational setting considered here, the requirements are specified in terms of temporal sequences comprising randomly structured inputs. Because input signals are defined independently of any external meaning, they need not resemble real-world sensory data. This design limits the usefulness of pretraining and increases confidence that any observed learning arises from the input stream itself.

\begin{figure}
    \centering
    \includegraphics{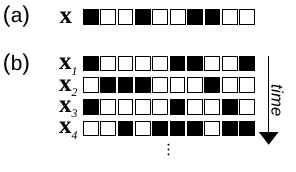}
    \caption{(a) Example of a 10-bit input $\mathbf x$ with four bits set. (b) Example of an input sequence.}
    \label{fig:input}
\end{figure}

Each input represents a snapshot of multiple binary signals at a single time step (Figure~\ref{fig:input}a). Spatial organisation within each input encodes local structure, whereas semantic richness arises from the temporal evolution of the input sequence (Figure~\ref{fig:input}b). The interaction between spatial and temporal dimensions gives rise to structured patterns that are challenging to adapt to.

\begin{figure}
    \centering
    \includegraphics{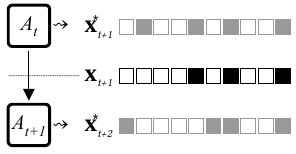}
    \caption{Iterative adaptation in discrete time. Following input $\mathbf x_t$, the system $A_{t}$ issues the prediction $\mathbf x^{*}_{t+1}$. Upon observing the realised input $\mathbf x_{t+1}$, it updates in response to the error in the second bit and produces the next one-step-ahead prediction $\mathbf x_{t+2}^{*}$.}
    \label{fig:model}
\end{figure}

An AGI system must learn from a stream of inputs. Upon receiving an input at time step $t$, it is required to issue a prediction $x_{t+1}^*$ for the subsequent input, as shown in Figure~\ref{fig:model}. A transition from configuration $A_t$ to $A_{t+1}$ is triggered by the arrival of the actual input ($x_{t+1}$). The central challenge is not simply extrapolation but discerning the underlying causes or regularities that produce the observed input stream and using that understanding to make accurate future predictions. 

\section{The hypothesis}

We hypothesise that any system capable of human-level general intelligence must satisfy at least the twelve requirements defined later. These requirements characterise fundamental properties of adaptive systems that learn from temporally structured inputs without relying on externally supplied semantic knowledge. The selection of these requirements is guided by observations of biological information processing and by the need to support autonomous learning from sensory signals.

The hypothesis is empirically falsifiable. It would be rejected by the existence of a demonstrably non-intelligent system that fulfils all requirements. The proposed requirements should therefore not be interpreted as a definition of intelligence.

In what follows, learning denotes the adaptive process through which a system becomes able to correctly anticipate future inputs based on its input history. This definition serves only to characterise prediction success as an operational capability, not an evaluation metric for intelligence. The term \emph{model} denotes a functioning instance of a system type that updates its internal configuration and produces a prediction of a future input based on its input history. Independent models share the same operating principles but are otherwise isolated from one another. 

\section{Preliminaries}

Let $\mathbf{x} \in X = \{0,1\}^{L}$ with $L = 10$ denote a ten-bit input vector, and let $\boldsymbol{\chi} = (\mathbf{x}_t)_{t=1}^N, \mathbf{x}_t\in X$ denote an input sequence with $N$ elements. For any input vector $\mathbf{x}\in X$, we write $\mathbf{x}[i]$ for its $i$-th bit.

Let $\mathcal{M}$ denote the set of all reachable\footnote{Starting from the initial configuration under some input history.} configurations of a system under AGITB evaluation. An individual system instance is denoted by $A \in \mathcal{M}$, with $B \in \mathcal{M}$ representing an independent instance of the same type. When temporal indexing is required, we write $A_t$ for the configuration obtained after $t$ update steps along its learning trajectory, with $A_0$ representing the initial configuration at the onset of learning. 

For brevity, we refer to an instance at a given internal configuration simply as a model.

\begin{definition}[Prediction]
A model’s prediction is given by the operator $\leadsto \subseteq\ \mathcal{M} \times X$, where
\[
    A_t \leadsto \mathbf{x}_{t+1}^*
\]
denotes that the model configuration $A_t \in \mathcal{M}$ predicts the next input to be $\mathbf{x}_{t+1}^* \in X$.
\end{definition}

\begin{definition}[Model update]
Model update is given by the transition relation $\xmapsto{\cdot}\ \subseteq\ \mathcal{M} \times X \times \mathcal{M}$, where
\[
A_t \xmapsto{\mathbf{x}_{t+1}} A_{t+1}
\]
denotes the atomic transition of the model configuration $A_t \in \mathcal{M}$ upon receiving the input $\mathbf{x}_{t+1} \in X$.
\end{definition}
Operationally, prediction and update are tightly coupled: processing the input $\mathbf{x}_{t+1}$ both updates the model and produces a prediction of the subsequent input $\mathbf{x}_{t+2}$. This allows us to write the combined transition as
\[
    A_t(\mathbf {x}_{t+1}) \rightarrow
    \bigl(A_{t+1},\, \mathbf{x}^*_{t+2}\bigr),
\]
which subsumes both model advancement and prediction.

\begin{definition}[Model trajectory]
For any model configuration $A \in \mathcal{M}$ and any input sequence
$\boldsymbol{\chi} = (\mathbf{x}_i)_{i=1}^k$, we write
\[
A \xmapsto{\;\boldsymbol{\chi}\;} A^{\boldsymbol{\chi}}
\]
to denote the transitions to the configuration obtained by sequentially updating the model with all inputs in $\boldsymbol{\chi}$, i.e.,
\[
A^{\boldsymbol{\chi}}
=\;
A^{(\mathbf{x}_1,\cdots,\mathbf{x}_k)}
:=\;
(\cdots((A \xmapsto{\mathbf{x}_1}) \xmapsto{\mathbf{x}_2})\cdots)\xmapsto{\mathbf{x}_k}.
\]
\end{definition}
Accordingly, $A_t^{\boldsymbol{\chi}}$ denotes the configuration obtained by applying $\boldsymbol{\chi}$ to $A_t$, that is, $A_t^{\boldsymbol{\chi}}:=(A_t)^{\boldsymbol{\chi}}$. A model may be exposed to multiple input sequences in succession; for two sequences $\boldsymbol{\chi}_1$ and $\boldsymbol{\chi}_2$, their concatenation is written $\boldsymbol{\chi}_1\boldsymbol{\chi}_2$, and the resulting configuration satisfies
\[
    A \xmapsto{\;\boldsymbol{\chi}_1\;} A^{\boldsymbol{\chi}_1}
       \xmapsto{\;\boldsymbol{\chi}_2\;} A^{\boldsymbol{\chi}_1\boldsymbol{\chi}_2}.
\]

\begin{definition}[Autoregressive generation]
Autoregressive generation is given by the relation $\Rrightarrow\ \subseteq\ \mathcal{M} \times \bigl(X^{*}\cup X^{\mathbb{N}}\bigr)$, where
\[
A_t \Rrightarrow \boldsymbol{\chi}^*
\]
denotes that the model configuration $A_t$ generates the (finite or infinite) sequence
$\boldsymbol{\chi}^* = (\mathbf{x}_{t+1}^*, \mathbf{x}_{t+2}^*, \ldots)$ by recursively feeding each predicted input back into the model; for a finite sequence, $A_t \Rrightarrow \boldsymbol{\chi}^*$ means that the first $|\boldsymbol{\chi}^*|$ generated inputs equal $\boldsymbol{\chi}^*$. In particular, for all $k \ge 0$ (and $k < |\boldsymbol{\chi}^*|$ for a finite sequence),
\[
A_{t+k} \leadsto \mathbf{x}_{t+k+1}^*,
\quad
A_{t+k} \xmapsto{\mathbf{x}_{t+k+1}^*} A_{t+k+1}.
\]
\end{definition}

\begin{definition}[Behavioural equivalence]
Two model configurations $A, B\in\mathcal{M}$ are \emph{behaviourally equivalent}, written $A \equiv B$, if they issue identical predictions under every subsequent input sequence:
\[
A \equiv B
\;:\!\iff\;
\forall\,\boldsymbol{\chi}\in X^{*},\ \forall\,\mathbf{x}\in X:\;
A^{\boldsymbol{\chi}} \leadsto \mathbf{x}
\iff
B^{\boldsymbol{\chi}} \leadsto \mathbf{x},
\]
where $X^{*}$ includes the empty sequence. Identical configurations are behaviourally equivalent; the converse need not hold.
\end{definition}

\begin{definition}[Learning]\label{def:learning}
The model $A$ is said to \emph{learn} a sequence
$\boldsymbol{\chi}$, written \(A \rhd \boldsymbol{\chi}\),
if there exists a finite number of passes over \(\boldsymbol{\chi}\), possibly none, after which \(A\) can
autoregressively reproduce a single instance of \(\boldsymbol{\chi}\), i.e.
\[
A \rhd \boldsymbol{\chi}
\;:\!\iff\;
\exists\, n \in \mathbb{N}_{\ge 0}:\;
A^{\boldsymbol{\chi}^{\,n}} \Rrightarrow \boldsymbol{\chi}.
\]
We write \(m_A(\boldsymbol{\chi}) \coloneqq \min\{n\ge0\mid A^{\boldsymbol{\chi}^n}\Rrightarrow\boldsymbol{\chi}\}\) for the number of passes that precede the first perfectly predicted one.
\end{definition}

\begin{definition}[Post-learning configuration]
If \(A \rhd \boldsymbol{\chi}\), the post-learning configuration is the configuration reached immediately after the first perfectly predicted pass:
\[
A_{\boldsymbol{\chi}} \coloneqq A^{\boldsymbol{\chi}^{\,m_A(\boldsymbol{\chi})+1}}.
\]
\end{definition}

\begin{definition}[Learning time]
The \emph{learning time in atomic steps} is defined as
\[
\tau_A(\boldsymbol{\chi}) \coloneqq
\begin{cases}
|\boldsymbol{\chi}|\cdot m_A(\boldsymbol{\chi}),
& \text{if } A\rhd\boldsymbol{\chi},\\
\infty,&\text{else}.
\end{cases}
\]

\end{definition}
Learning time counts the atomic steps that precede the first perfectly predicted pass; it is zero if the model already predicts the sequence. It measures the first occurrence of accurate prediction and does not presuppose permanent retention: the post-learning configuration need not reproduce the sequence again. If the model never learns to predict the sequence, its learning time is infinite.

\section{The requirements}

\begin{req}[Uninformed start]\label{R:uninformed}
All models begin learning from the identical initial configuration $\diamond$.
\[
\forall\, A:\;
A_0 = \diamond.
\]
\end{req}

The initial model is assumed to contain no environment-specific knowledge beyond the architectural biases necessary for learning. All knowledge relevant to input must be acquired solely through subsequent exposure to data. This assumption is commonly regarded as a necessary condition for universality, defined as the ability to adapt to arbitrary environments. The formal statement captures only the identical starting configuration; the absence of environment-specific knowledge is an assumption that the requirement expresses but that no test can verify, since comparing fresh instances of a model cannot reveal knowledge that all of them share.

We assume that general-purpose learning systems, including biological brains, do not begin with an innate understanding of external inputs but instead acquire meaning through interaction with their environment. Each system must construct semantic content from raw sensory data rather than rely on pre-encoded knowledge.\footnote{Although reflexes may be genetically specified, they do not constitute genuine understanding. They are rather evolutionary features of the subcortical ``old'' brain and not prerequisites for intelligence \cite[p.~66]{Hawkins2005}.}

\begin{req}[Determinism]\label{R:determ}
Model evolution is deterministic with respect to input.
\[
\forall A,\ \forall\,\mathbf{x} \in X:\;
\exists!\,A'
\text{ such that }
A \xmapsto{\mathbf{x}} A'.
\]
\end{req}

Biological neurons operate in a functionally deterministic manner, ensuring stability and consistency in brain function. Although minor stochastic effects may occur, they do not undermine the rule-governed nature of neural processing. By analogy, we assume that input history uniquely determines the model configuration. A model is therefore fully determined by its input history.

Determinism at the level of neural signal processing is necessary for stable, reproducible brain function. In contrast, the apparent unpredictability of cognition stems from the system's complexity rather than from genuine indeterminacy \cite{Cave2016}.

\begin{req}[Trace]\label{R:trace}
Each input leaves a permanent internal trace.
\[
\forall\,\boldsymbol{\chi}_1\neq\boldsymbol{\chi}_2\in X^{*},\
\forall\,\boldsymbol{\chi}\in X^{*}:\
A_0^{\boldsymbol{\chi}_1\boldsymbol{\chi}} \not\equiv A_0^{\boldsymbol{\chi}_2\boldsymbol{\chi}}.
\]
\end{req}

Two models whose input histories differ never become indistinguishable, whatever they experience afterwards: every input leaves a trace that no later input can erase, and that eventually shows in behaviour. It follows that configurations never repeat and trajectories contain no cycles (Corollary~\ref{R:transition}).

In the reference implementation, two models start one input apart: one receives an input twice in a row, violating the refractory constraint of Req.~\ref{R:arp}, while the other receives it followed by silence. Both then live the same long life, in which both beginnings recur repeatedly, so that nothing either model experienced is unique to it. Afterwards they must issue different predictions somewhere on a long random sequence of probe inputs. A model that only records which inputs it has seen, and what followed them, cannot tell the two apart; only a trace of the history itself, including the order and multiplicity of its events, survives the shared life.

Human brains satisfy an effective version of this requirement at the lowest physical level. Each sensory or internal event induces irreversible microstate changes that are never exactly revisited. Through complex internal interactions, this persistent history contributes to the apparent unpredictability of decision-making. Although information may appear discarded, compressed, or behaviourally inaccessible at the cognitive level, we posit that internal model configurations continue to develop in a distinct manner over time, consistent with the formation of permanent internal traces.

Taken literally, the requirement is an idealisation. Since distinct histories must lead to distinguishable configurations, a model satisfying it exactly needs infinitely many configurations, and its memory must grow with the length of its history; no system with a fixed amount of memory can satisfy it exactly. Like the other requirements that quantify over unbounded domains, it is therefore approximated: the test checks it on a single hard case over a finite horizon, and passing it means that no forgetting was detected within that horizon, not that the trace is permanent in the strict sense.

\begin{req}[Time]\label{R:time}
Model evolution depends on input order.
\[
    \forall\,A, \forall\,\boldsymbol{\chi}_1,\boldsymbol{\chi}_2\in X^+ :\;
    \boldsymbol{\chi}_1\boldsymbol{\chi}_2\neq\boldsymbol{\chi}_2\boldsymbol{\chi}_1
    \implies
    A^{\boldsymbol{\chi}_1\boldsymbol{\chi}_2} \neq A^{\boldsymbol{\chi}_2\boldsymbol{\chi}_1}.
\]
\end{req}

Model evolution depends intrinsically on input order: for any system, changing the input order necessarily results in a novel configuration. This enforces a strict temporal asymmetry in learning dynamics and rules out commutative or order-invariant update mechanisms. Sensitivity to temporal structure in this strong sense is regarded as a defining property of intelligent systems.

Formally, this requirement is implied by Req.~\ref{R:trace}: the histories $\boldsymbol{\chi}_1\boldsymbol{\chi}_2$ and $\boldsymbol{\chi}_2\boldsymbol{\chi}_1$ differ, so the resulting configurations are not even behaviourally equivalent. It is nevertheless stated separately, because order sensitivity is a central property in its own right and admits a simple, direct test: in the reference implementation, two copies of a model with a random input history receive the same two inputs in opposite order and must end in different configurations.

\begin{definition}[Admissible sequences]
The set of \emph{admissible} sequences is defined as
\[
\boldsymbol{\Phi}
:=\Bigl\{
(\mathbf x_1,\dots,\mathbf x_k)\in X^k\ \Big|\ 
k\geq 1
\]
\[ 
\wedge\
\forall i\in\{1,\dots,k\}:\ 
\mathbf{x}_i\cdot\mathbf{x}_{((i\bmod k)+1)} = 0
\Bigr\}.
\]
\end{definition}

\begin{req}[Absolute refractory period]\label{R:arp}
A model can learn a cyclic sequence only if successive inputs, including the last and the first, share no active bit.
\[
\forall\,\boldsymbol{\chi}:\;
A_0 \rhd \boldsymbol{\chi} \;\implies\; \boldsymbol{\chi}\in\boldsymbol{\Phi}.
\]
\end{req}

Biological intelligence relies on discrete spiking events for communication and learning, and individual neurons cannot fire again immediately after activation. The absolute refractory-period constraint defines admissible sequences but does not imply that all such sequences are learnable.

Absolute refractory periods impose a minimum separation between spikes, creating temporal structure that supports temporally sensitive learning processes \cite{Gerstner2002}. All temporal sequences that satisfy biologically plausible refractory constraints are therefore admissible, independent of any semantic interpretation of the signals.

\begin{definition}[Learnable sequences]
The set of \textit{learnable} sequences is the subset
\[
\boldsymbol{\Psi} := \bigl\{ \boldsymbol{\phi}\in\boldsymbol{\Phi}\ \big|\ 
A_0 \rhd \boldsymbol{\phi}\bigr\}.
\]
\end{definition}

\begin{definition}[Learning in turn]
Sequences $\boldsymbol{\psi}_1,\ldots,\boldsymbol{\psi}_k$ are \emph{learned in turn} by a configuration $A$ if each of them is learned by the configuration that results from learning the previous ones:
\[
C_0 = A,\qquad
C_{i-1} \rhd \boldsymbol{\psi}_i,\qquad
C_i = (C_{i-1})_{\boldsymbol{\psi}_i},\qquad
i = 1,\ldots,k.
\]
We write $A_{\boldsymbol{\psi}_1\cdots\boldsymbol{\psi}_k} := C_k$, with $A_{\boldsymbol{\psi}_1\cdots\boldsymbol{\psi}_k} = A$ for $k=0$.
\end{definition}

\begin{req}[Inevitable saturation]\label{R:saturation}{\ }\\
(a) {A model cannot learn everything there is to learn.}
\[
\forall\,A\in\mathcal{M},\;
\exists\,k\ge 0,\;
\exists\,\boldsymbol{\psi}_1,\ldots,\boldsymbol{\psi}_k \in \boldsymbol{\Psi} \text{ learned in turn by } A,
\]
\[
\exists\,\boldsymbol{\psi}\in\boldsymbol{\Psi}:\;
\neg\bigl(
A_{\boldsymbol{\psi}_1\cdots\boldsymbol{\psi}_k}
\rhd \boldsymbol{\psi}
\bigr).
\]
(b) {All admissible length-2 sequences are universally learnable.}
\[
\forall\,A,\ \forall\,\boldsymbol{\phi}\in\boldsymbol{\Phi}\cap X^2:\ A\rhd\boldsymbol{\phi}.
\]
\end{req}

Learning systems with finite representational and adaptive capacity necessarily face limits on the extent to which they can learn sequences of length greater than two, whereas all admissible sequences of length two are universally learnable. Saturation must be reachable from every configuration, though not along every learning path: a model that learns only length-two sequences never saturates, by Req.~\ref{R:saturation}(b). The sequences that a saturated model can no longer learn are therefore longer than two.

Length-two sequences occupy a special position, as they are the shortest sequences that encode temporal relationships, while a single input carries no temporal information. They define transitions that are the irreducible units of temporal prediction and thus necessary for temporal learning. 

A related distinction is observed in biological learning systems. Humans can readily acquire arbitrary pairwise associations, whereas learning and retaining longer temporal structures is subject to pronounced capacity limits and interference effects.

In the reference implementation, a model with a random input history keeps learning random cyclic sequences of seven inputs until it meets one that it cannot learn although a fresh model can; it must meet such a sequence eventually, and must afterwards still learn a random admissible sequence of length two.

\begin{req}[Temporal adaptability]\label{R:temporal}
The model must be able to learn sequences with varying cycle lengths and adapt when the temporal scale changes.
\[
\exists\,\boldsymbol{\psi}_1 \in \boldsymbol{\Psi},\ \exists\,\boldsymbol{\psi}_2 \in \boldsymbol{\Phi}:\ 0<|\boldsymbol{\psi}_1|<|\boldsymbol{\psi}_2| \ \wedge\ (A_0)_{\boldsymbol{\psi}_1} \rhd \boldsymbol{\psi}_2.
\]
\end{req}

This requirement captures the ability to learn and track temporal structure across multiple timescales. Systems restricted to rigid pattern matching at a single, fixed periodicity are therefore excluded, as adaptive learning requires sensitivity to recurring structure at different temporal scales.

In the reference implementation, a fresh model must learn a rhythm of silent inputs in which a single bit spikes every seventh step, and then the same rhythm with a spike every eighth step. Predicting the spike requires counting the silent steps that precede it, and the count changes between the two sequences.

\begin{req}[Content sensitivity]\label{R:content}
Adaptation time is input-dependent.
\[
\exists\,\boldsymbol{\psi}_1,\,\boldsymbol{\psi}_2\in \boldsymbol{\Psi}:\ 
|\boldsymbol{\psi}_1| = |\boldsymbol{\psi}_2|
\;\wedge\;
\tau_{A_0}(\boldsymbol{\psi}_1)
\neq
\tau_{A_0}(\boldsymbol{\psi}_2).
\]
\end{req}

The rate at which a model adapts depends on the structure of the input sequence. Simple or highly regular sequences typically yield rapid convergence, whereas less regular inputs require longer exposure before the model can reliably capture the underlying structure.

\begin{req}[Context sensitivity]\label{R:context}
Adaptation time is model-dependent.
\[
\exists\,\boldsymbol{\psi} \in \boldsymbol{\Psi},\; \exists\, A \neq B:\quad
\tau_A(\boldsymbol{\psi}) < \infty
\;\wedge\;
\tau_B(\boldsymbol{\psi}) < \infty
\;\wedge\;
\tau_A(\boldsymbol{\psi}) \neq \tau_B(\boldsymbol{\psi}).
\]
\end{req}

A model's internal configuration reflects the cumulative influence of past inputs and thereby determines the context in which new information is learned. When subsequent inputs are consistent with the structure established through prior learning, adaptation may proceed rapidly. Conversely, when new inputs conflict with this learned context, additional iterations may be required before accurate prediction becomes possible.

\begin{req}[Denoising]\label{R:denoise}
An informed model consistently outperforms its uninformed self at predicting what follows a corrupted input.

Let $A$ be an adult model, that is, a model that has reached the saturation of Req.~\ref{R:saturation}(a). Let $\boldsymbol{\phi}=(\mathbf{x}_1,\dots,\mathbf{x}_k)\in\boldsymbol{\Phi}$ be a random cyclic sequence that the adult learns, $A\rhd\boldsymbol{\phi}$, and let
$\boldsymbol{\phi}'=(\mathbf{x}_1,\dots,\mathbf{x}_{k-1},\mathbf{x}_k')$ be obtained from
$\boldsymbol{\phi}$ by flipping a single bit of its last input, such that $\boldsymbol{\phi}'\in\boldsymbol{\Phi}$. The informed model learns $\boldsymbol{\phi}$ and then observes $\boldsymbol{\phi}'$, whereas the uninformed model observes only $\boldsymbol{\phi}'$:
\[
A_{\boldsymbol{\phi}}^{\boldsymbol{\phi}'} \leadsto \mathbf{x}^{*}_{\mathrm{inf}},
\qquad
A^{\boldsymbol{\phi}'} \leadsto \mathbf{x}^{*}_{\mathrm{uninf}}.
\]
Over $T$ independently generated pairs $(\boldsymbol{\phi}^{(i)},\boldsymbol{\phi}'^{(i)})$, the informed model predicts the input that follows the familiar sequence exactly more often than the uninformed one:
\[
\sum_{i=1}^{T}\mathbf{1}\bigl\{\mathbf{x}^{*(i)}_{\mathrm{inf}}=\mathbf{x}^{(i)}_1\bigr\}
>
\sum_{i=1}^{T}\mathbf{1}\bigl\{\mathbf{x}^{*(i)}_{\mathrm{uninf}}=\mathbf{x}^{(i)}_1\bigr\}.
\]
In the reference implementation, $T=20$ and $k=7$.
\end{req}

A model is required to recognise a familiar sequence despite a single corrupted input, and to predict what normally follows it. Because an adult is saturated, it cannot learn every sequence; only sequences it does learn can become familiar, so sequences it fails to learn are redrawn. The comparison is made against the same adult model without knowledge of the sequence, so that the advantage must come from the familiar sequence rather than from general experience or from a fortunate guess. Using adult models also ensures that denoising is demonstrated by a model whose capacity is already in use, rather than by a fresh model with an empty memory. Every adult lives the same life, which keeps failures reproducible.

\begin{req}[Generalisation]\label{R:general}
An informed model applies the rule governing its world to an input it has never seen, which an uninformed model cannot do.

Let $\boldsymbol{\psi}_1,\ldots,\boldsymbol{\psi}_k\in\boldsymbol{\Phi}$ be cases of a world: cyclic sequences whose changes follow a common rule, possibly with repetitions. Let $\mathbf{p}\in X$ be a prefix input that occurs in none of the cases, and let $\mathbf{c}\in X$ be the continuation that the rule assigns to $\mathbf{p}$. The informed model learns the cases in turn and then observes $\mathbf{p}$; the uninformed model observes only $\mathbf{p}$. The informed model must predict the continuation exactly, and the uninformed one must not:
\[
\bigl((A_0)_{\boldsymbol{\psi}_1\cdots\boldsymbol{\psi}_k}\bigr)^{(\mathbf{p})} \leadsto \mathbf{c},
\qquad
A_0^{(\mathbf{p})} \not\leadsto \mathbf{c}.
\]
In the reference implementation, each case shows a pattern that appears, moves one position to the right at each time step, and stops, leaving silence:
\[
\boldsymbol{\gamma}_1 =
\left(
\begin{array}{l}
\texttt{1000100001},\\
\texttt{0100010000},\\
\texttt{0010001000},\\
\texttt{0000000000}
\end{array}
\right),
\quad
\boldsymbol{\gamma}_2 =
\left(
\begin{array}{l}
\texttt{0000101000},\\
\texttt{0000010100},\\
\texttt{0000000000}
\end{array}
\right),
\]
\[
\boldsymbol{\gamma}_3 =
\left(
\begin{array}{l}
\texttt{1010000010},\\
\texttt{0101000001},\\
\texttt{0010100000},\\
\texttt{0001010000},\\
\texttt{0000101000},\\
\texttt{0000000000}
\end{array}
\right).
\]
They are learned in the order $(\boldsymbol{\psi}_1,\ldots,\boldsymbol{\psi}_5)=(\boldsymbol{\gamma}_1,\boldsymbol{\gamma}_2,\boldsymbol{\gamma}_1,\boldsymbol{\gamma}_3,\boldsymbol{\gamma}_2)$, with $\mathbf{p}=\texttt{0001001000}$ and $\mathbf{c}=\texttt{0000100100}$, written with bit~9 on the left.
\end{req}

Improving performance beyond memorisation requires generalisation: a model that has merely memorised its world has nothing to say about an input it has never seen, whereas a model that has grasped the rule governing the world can apply it. The informed model must learn every case of the world before it is asked to generalise. Because the cases are learned in an interleaved order, and the first two are revisited after others have been learned, what the model knows about one case must survive learning the others.

The prefix arrives after silence, where every case starts a new pattern that is then shifted; the rule therefore assigns the prefix its own shift as the continuation. The cases are constructed so that this continuation cannot be obtained by tracking which input channel follows which: because every pattern eventually stops, most channels are sometimes followed by nothing at all, and statistics of individual channels predict only part of the continuation. Only the movement of the input as a whole explains every transition. The prefix keeps both transitions admissible and differs from every input of the cases, silence included, in at least two bits, so that it cannot be mistaken for a noisy familiar input of the kind considered in Req.~\ref{R:denoise}.

The test samples the requirement with a single, deliberately simple world. The requirement concerns any rule a model can extract from its experience, whereas the test checks only one rule: the movement of a pattern across neighbouring channels. Passing the test therefore demonstrates generalisation of this rule only, and a model engineered with knowledge of the rule can pass it (Section~\ref{sec:exploitation}); conversely, a model that generalises other kinds of rules but ignores the arrangement of channels fails it. As with the other requirements, the test approximates an unbounded criterion with a finite, transparent instance.

\begin{req}[Real-time liveness]\label{R:liveness}
Each model update completes within a fixed wall-clock time bound, independent of the input history.
\[
\exists\,t_{\max}>0:\;
\Delta t(A,\mathbf{x})\le t_{\max}
\quad\forall A\in\mathcal M,\ \forall\mathbf x\in X,
\]
where $\Delta t(A,\mathbf{x})$ denotes the wall-clock time required to
perform the atomic transition
$A \xmapsto{\mathbf{x}} A'$.
\end{req}

A model must complete each atomic transition within a bounded amount of time to remain suitable for real-time interaction. Biological brains satisfy this requirement in large part through massive parallelism: configuration transitions and signal emissions occur concurrently across large neuronal populations, with neural signal propagation times bounded under normal operating conditions. As a result, cognitive processing does not exhibit unbounded slowing as a function of accumulated experience or instantaneous sensory load.

In the reference implementation, the time a model needs to process a batch of inputs is measured on a fresh model and again after a long random input history; the second measurement must remain below twice the first. Each measurement is the fastest of several timings on copies of the model, so that a single interruption by the operating system cannot decide the test.

\subsection{Immediate consequences of the requirements}

Two immediate consequences follow from the preceding requirements. First, any system satisfying Req.~\ref{R:trace} undergoes perpetual change: every input induces a distinct internal configuration, and model trajectories contain no exact recurrences. Second, Reqs.~\ref{R:trace} and~\ref{R:saturation}(b) together imply a form of behavioural unobservability, in which distinct internal configurations can yield identical autoregressive outputs.

\begin{corollary}[Perpetual change]\label{R:transition}
Configurations never repeat; in particular, every input induces a change in the model's internal configuration.
\[
\forall\, t\neq s\ge 0:\;
A_t\neq A_s,
\qquad
\forall\, t\ge 0:\;
A_{t+1}\neq A_t.
\]
\end{corollary}
\begin{proof}
Let $A_t = A_0^{\boldsymbol{\chi}_t}$ and $A_s = A_0^{\boldsymbol{\chi}_s}$ with $t\neq s$, so that the histories $\boldsymbol{\chi}_t$ and $\boldsymbol{\chi}_s$ differ in length. If $A_t = A_s$, the two configurations would evolve identically under every continuation and would therefore be behaviourally equivalent, contradicting Requirement~\ref{R:trace}. Setting $s=t+1$ yields $A_{t+1} \neq A_t$ for all $t$.
\end{proof}

\begin{corollary}[Unobservability]\label{cor:unobserve}
Distinct model configurations may be observationally indistinguishable under autoregressive generation:
\[
\exists\,\boldsymbol{\phi}\in\boldsymbol{\Phi}\;
\exists\,A \neq B:\;
A \Rrightarrow \boldsymbol{\phi}
\;\wedge\;
B \Rrightarrow \boldsymbol{\phi}.
\]
\end{corollary}
\begin{proof}
Choose \(\boldsymbol{\phi}\in\boldsymbol{\Phi}\cap X^2\). By
Requirement~\ref{R:saturation}(b), every reachable configuration learns
\(\boldsymbol{\phi}\).

Let \(A\) and \(B\) be two reachable configurations whose input histories
differ but have equal length. Since both learn \(\boldsymbol{\phi}\), the
configurations
\(A' = A^{\boldsymbol{\phi}^{\,m_A(\boldsymbol{\phi})}}\) and
\(B' = B^{\boldsymbol{\phi}^{\,m_B(\boldsymbol{\phi})}}\), which precede their
first perfectly predicted passes, satisfy
\[
A'\Rrightarrow\boldsymbol{\phi},
\qquad
B'\Rrightarrow\boldsymbol{\phi}.
\]
Their histories extend those of \(A\) and \(B\) by whole passes of
\(\boldsymbol{\phi}\): if the numbers of passes are equal, the extended
histories still differ where the original ones did; otherwise they differ in
length. By Requirement~\ref{R:trace}, configurations with different histories
are behaviourally distinguishable, and thus distinct, so \(A'\neq B'\).
\end{proof}

\section{Methodology}

The proposed requirements are operationalised as concrete evaluation procedures, with each requirement evaluated by an empirical test whose failure indicates non-compliance. The objective is not to demonstrate intelligence, but to expose violations under stringent conditions, thereby probing the robustness of candidate systems. 

Taken together, the requirement-specific tests form a practical testbed, the Artificial General Intelligence Testbed (AGITB), that cannot, and is not intended to, prove that a model satisfies all requirements. Rather, its strength lies in identifying violations of individual requirements and in indicating which constraints are breached. A reference C++ implementation of AGITB is freely available under the GPL license at \url{https://github.com/matejsprogar/agitb}.

The reference implementation of AGITB was released publicly as free software in April 2025, together with a preprint describing the testbed, to encourage independent attempts at satisfying the requirements. No independently verified solution satisfying all tests has been reported.

\subsection{Inherent limitations}

AGITB confronts three principal challenges. First, several requirements quantify over effectively unbounded sets of cases and cannot be exhaustively verified through finite testing. Second, individual requirements cannot always be captured by a single, easily interpretable test. Third, external performance metrics are unsuitable, as they cannot reliably distinguish generalisable adaptive behaviour from task-specific optimisation.

We address the first challenge by approximating unbounded domains with finite, tractable collections of test cases that provide empirical support for a given claim. The second is addressed through a synergistic test design in which multiple simple tests are mutually informative and reinforce one another. In such settings, failure to satisfy a requirement may not be revealed by a test in isolation, but instead emerges when constraints imposed by other tests become mutually incompatible. The third challenge is addressed by constructing controlled scenarios involving independent system instances and assessing behaviour through systematic comparison, with success defined in terms of configuration or behavioural distinctions rather than external task scores.

\subsection{Test design}

In the reference implementation, a compromise approximation for unlimited iterations is set to 5{,}000. While finite, this value was chosen to balance computational feasibility with stringency and is used consistently across tests. To minimise the influence of stochastic effects, every test that involves random inputs is repeated 5,000 times, and the system under evaluation must succeed on each trial; the three tests without random inputs (Reqs.~\ref{R:uninformed}, \ref{R:temporal}, and~\ref{R:general}) are run once. The resulting procedure functions as an extreme form of stress testing, ensuring that passing results reflect systematic robustness rather than chance. 

The tests are not designed as conventional significance tests, nor to maximise statistical power; instead, they prioritise reliability by admitting virtually no noise or marginal effects. Modest or inconsistent improvements are treated as failures.

Finally, AGITB adopts an all-or-nothing evaluation criterion: a system under evaluation must satisfy all requirement-specific tests to pass the benchmark. While individual tests may be solvable in isolation, the tests are designed such that simultaneous satisfaction of all requirements imposes substantially stronger constraints.

\subsection{Problem space}

To prevent models from relying on brute-force memorisation, the problem space must be sufficiently large to exceed the capacity of any model operating under realistic computational constraints on both time and memory. In AGITB, tasks typically require predicting a temporal sequence of seven inputs ($N = 7$), each consisting of ten bits ($L = 10$). The resulting problem space is $\mathcal{S}=(\{0,1\}^{L})^N$, which has cardinality $|\mathcal{S}| = 2^{70}$, representing all possible binary input sequences of that length.

The \emph{absolute refractory period} prohibits any neuron (bit) from firing in consecutive time steps. This requirement substantially lowers the number of admissible sequences, yielding a subset $\mathcal{S}'\subset\mathcal{S}$. There are $|\mathcal{S}' | = (\mathcal{F}_{N+2})^{L} = 34^{10} \approx 2^{51}$ distinct seven-step temporal sequences of ten bits under the condition that a $1$ never carries over to the next time step, where $\mathcal{F}_i$ denotes the $i$-th Fibonacci number with $\mathcal{F}_0 = 0$.

In some cases, AGITB further constrains $\mathcal{S}'$ by requiring sequences to be cyclic, such that the first input also satisfies the absolute refractory constraint with respect to the final input in the sequence. The resulting set of cyclic temporal sequences respecting the refractory constraint, $\mathcal{S}''=\boldsymbol{\Phi}\cap X^{N}\subset \mathcal{S}'$, has cardinality $|\mathcal{S}'' | = (\mathcal{L}_{N})^{L} = 29^{10} \approx 2^{49}$, where $\mathcal{L}_i$ denotes the $i$-th Lucas number with $\mathcal{L}_0 = 2$.

The choice of seven-step sequences with ten-bit inputs is sufficient to expose the naive and specialised systems we have evaluated while maintaining computational efficiency; it does not, however, prevent systems engineered around the tests from passing (Section~\ref{sec:exploitation}). Increasing these default values could exceed the capabilities of the first-generation AGI under evaluation, potentially leading to false negatives and substantially increasing runtime. The current configuration, therefore, ensures that each test remains both computationally feasible and diagnostically informative.

\subsection{Benchmark Exploitation}\label{sec:exploitation}

Because AGITB consists of intentionally simple and transparent tests, a model can be engineered to optimise for the benchmark rather than for the underlying learning capabilities it is intended to measure.

For example, a model with a built-in mechanism that shifts its input along the channels can pass the test of Req.~\ref{R:general} without learning the rule from the world it is presented with, even though such pre-installed knowledge is prohibited by Req~\ref{R:uninformed}. One might argue that the shift mechanism is merely an architectural choice. However, in this case, it is precisely the rule on which the test is based, and any other operation used by the test could be built into the model in the same way. Such a model is of little use in general, as it does not learn the rules governing the world but brings them along. Unfortunately, no test of Req.~\ref{R:uninformed} can detect pre-installed knowledge, because it can only compare fresh instances of a model with one another, and all instances share the same built-in knowledge.

Other avenues include exploiting weaknesses in the benchmark's randomisation or exploiting properties of the comparison procedure between model instances. Such strategies optimise performance with respect to the benchmark implementation rather than the underlying capabilities being assessed.

Benchmark-specific design need not be as blatant as a single built-in rule. In our own attempts to break the testbed, we assembled a learner from generic components: an online-trained neural network, statistics of which input channels follow which, a switch that falls back on these statistics when the network's prediction is contradicted, and a built-in preference for the shortest shift of the whole input that explains a transition. With its parameters tuned against the failures observed during development, this learner passed all twelve tests. None of its components refers to a test explicitly, yet the system as a whole was shaped by knowledge of the tests, and it is not intelligent in any meaningful sense. AGITB therefore cannot be regarded as foolproof against fake solutions: it rejects systems whose violations its tests probe, but a sufficiently informed designer can pass every test without satisfying the requirements the tests approximate.

In principle, the benchmark could be hardened against these forms of exploitation by introducing additional randomisation, obfuscation, or verification mechanisms. However, these measures would inevitably increase implementation complexity while reducing transparency, interpretability, and reproducibility. Since AGITB is intended not only as an evaluation protocol but also as a practical development tool, the reference implementation deliberately favours simplicity and readability over resistance to benchmark-specific optimisations.

A more fundamental limitation arises from the need to approximate inherently unbounded evaluation criteria using finite computational resources. Several AGITB requirements would ideally be evaluated over an unbounded number of test cases, which is computationally infeasible. Instead, the benchmark performs a fixed number of iterations intended to approximate an otherwise indefinite evaluation process. In the reference implementation, this limit is controlled by the parameter \texttt{SimulatedInfinity}, which is currently set to 5{,}000.

Although this value is necessarily finite, our experiments indicate that, in combination with the current benchmark configuration (temporal patterns consisting of seven 10-bit input signals), it is sufficient to expose the violations of the systems we have evaluated that were not engineered around the tests, while maintaining computational efficiency suitable for rapid iterative experimentation.

The AGITB reference implementation is therefore intentionally designed to be transparent, computationally efficient, and straightforward to inspect. Implementation-specific exploits are not hypothetical, but hardening the benchmark against them would require secrecy or obfuscation that conflict with its purpose, and would still not make it foolproof; increasing its complexity for this reason would diminish its value as an interpretable, reusable and reproducible evaluation framework.

\subsection{Expected behaviour of representative system classes}

To contextualise the benchmark, we consider how representative classes of systems would be expected to behave under AGITB. These observations are qualitative reference points, not formal verification.

Human cognition provides a conceptual baseline. The requirements cannot be verified in humans directly, because cortical configurations cannot be inspected or compared, but many of them correspond to well-known cognitive abilities, such as temporal pattern recognition, denoising of corrupted inputs, and knowledge saturation. The notable exception is the uninformed start: adult humans carry accumulated prior knowledge, so the condition plausibly holds only at the onset of cortical development.

Classical symbolic programs encode prior assumptions about the meaning of the signals they process; their structure is knowledge supplied by the designer. Individual tests may be solvable by such specialised procedures, but the embedded knowledge contradicts the assumption behind the uninformed start, even though identical copies of a program satisfy its formal statement.

Artificial neural networks show the same distinction. Pretrained networks acquire their parameters from structured datasets in a separate phase driven by an external optimisation process, which conflicts with the assumption behind the uninformed start. Large language models are a particular case: they violate this assumption by construction, although identical copies satisfy its formal statement and its test, and their fixed-length context windows may break the uninterrupted trace of past inputs (Req.~\ref{R:trace}) once earlier tokens are truncated or compressed.

Networks trained online avoid these objections: they start from a fixed configuration and adapt continuously, with one update per input and the error in predicting that input as their only learning signal. They fit the setting of AGITB naturally, and some requirements hold largely by construction: a fixed architecture updates in constant time (Req.~\ref{R:liveness}), and continuous weights retain traces of past inputs that later shared experience does not erase (Req.~\ref{R:trace}). In our experiments, however, a small network of this kind violated other requirements. After a long random history, it could lose the ability to learn simple length-two sequences (Req.~\ref{R:saturation}(b)) unless its learning rate and weights were constrained, and it did not generalise the rule of Req.~\ref{R:general}, because associating windows of past inputs with their successors does not reveal how the input as a whole moves.

Individual tests may thus be solvable by specialised systems, but satisfying all requirements simultaneously imposes a substantially stronger constraint on candidate architectures.

\section{Conclusion}

Unlike conventional approaches to defining intelligence through high-level behaviours such as question answering or language translation, this work focuses on low-level, biologically grounded computational properties. The proposed requirements are tightly interdependent: no single requirement is decisive in isolation, yet their complementarity permits each to remain atomic in scope while jointly demanding the adaptive learning capabilities expected of an AGI.

The proposed requirements are not intended as a sufficient criterion for artificial general intelligence, nor do they characterise intelligence in any comprehensive sense. A system that satisfies them does not thereby exhibit higher-level capacities such as reasoning, abstraction, or natural language competence. Instead, the requirements target a set of low-level capabilities that may serve as necessary precursors to more general forms of intelligence, without themselves guaranteeing their emergence.

The central claim advanced here cannot be verified until a precise, universally accepted definition of intelligence is available, a prospect that is unlikely. It is, however, empirically falsifiable: the existence of a narrow system that satisfies all requirements would invalidate the claim that these properties are necessary for general intelligence. Passing the testbed is not the same as satisfying the requirements: the tests only approximate the requirements, and some of them, notably the uninformed start (Req.~\ref{R:uninformed}), cannot be verified by any test. A system engineered around the tests that passes them therefore exposes a limitation of the testbed rather than falsifying the hypothesis.

To date, no artificial system known to the authors has been shown to satisfy all requirements simultaneously without relying on knowledge of the tests, despite each requirement being well-defined and individually attainable. Because systems engineered around the testbed can pass it, passing the AGITB is not by itself evidence of crossing the boundary the requirements delineate; that boundary may nevertheless represent a pathway toward more general forms of intelligence.

\section*{Code availability}
The reference implementation of the proposed testbed is freely available at \url{https://github.com/matejsprogar/agitb}.

\section*{Funding}
The author acknowledges the financial support from the Slovenian Research Agency (research core funding No. P2-0057).



\begin{IEEEbiography}[{\includegraphics[width=1in,height=1.25in,clip,keepaspectratio]{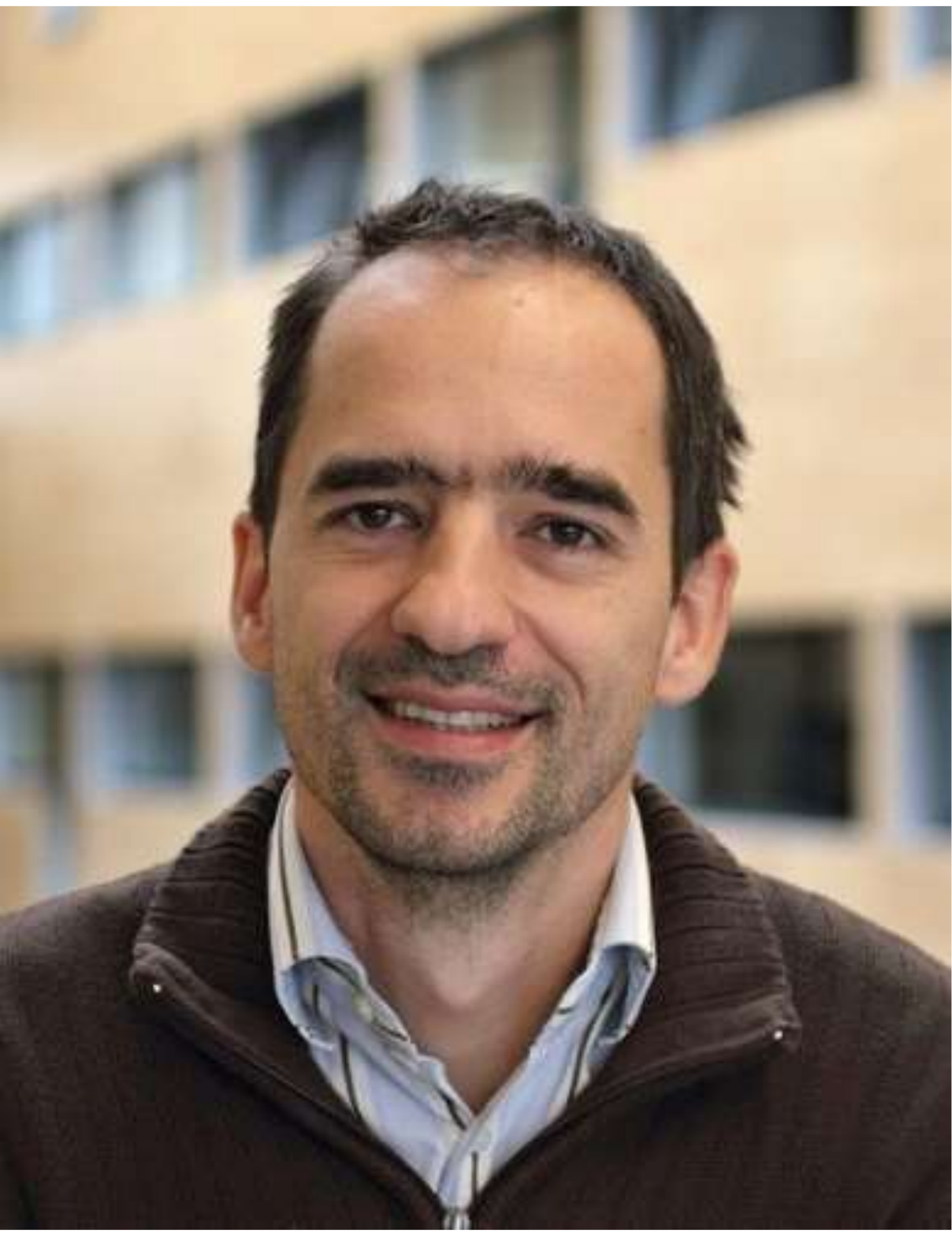}}]{Matej \v{S}progar} is an Assistant Professor at the University of Maribor, Faculty of Electrical Engineering and Computer Science, Maribor, Slovenia. He received the Ph.D. degree in computer science from the University of Maribor, where his doctoral research focused on autonomous evolutionary construction of decision trees. His research interests include artificial general intelligence, machine learning, and evolutionary computation. Contact him at matej.sprogar@um.si.
\end{IEEEbiography}

\end{document}